\pdfoutput=1

\documentclass[11pt]{article}

\usepackage{EACL2023}

\usepackage[T1]{fontenc}

\usepackage[utf8]{inputenc}

\usepackage{microtype}

\usepackage{inconsolata}

\usepackage{times}
\usepackage{latexsym}
\usepackage{amssymb}
\usepackage{multirow}
\usepackage{booktabs}
\usepackage{hyperref}
\usepackage{xspace}
\usepackage{amsmath}
\usepackage{CJKutf8}
\usepackage{xcolor}
\usepackage{graphicx}
\usepackage{caption}
\usepackage{subcaption}
\usepackage{duckuments}
\usepackage{arydshln}

\usepackage{bm}

\def\eqref#1{equation~\ref{#1}}

\def\1{\bm{1}}

\def\vu{{\bm{u}}}
\def\vv{{\bm{v}}}

\def\vz{{\bm{z}}}

\DeclareMathAlphabet{\mathsfit}{\encodingdefault}{\sfdefault}{m}{sl}
\SetMathAlphabet{\mathsfit}{bold}{\encodingdefault}{\sfdefault}{bx}{n}

\newcommand{\Ls}{\mathcal{L}}

\usepackage{array}

\definecolor{maroon}{RGB}{191, 96, 96}

\definecolor{purple}{RGB}{100, 0, 200}

\definecolor{mint}{rgb}{0.24, 0.71, 0.54}

\newcolumntype{L}[1]{>{\raggedright\let\newline\\\arraybackslash\hspace{0pt}}m{#1}}
\newcolumntype{C}[1]{>{\centering\let\newline\\\arraybackslash\hspace{0pt}}m{#1}}
\newcolumntype{R}[1]{>{\raggedleft\let\newline\\\arraybackslash\hspace{0pt}}m{#1}}

\makeatletter
\def\adl@drawiv#1#2#3{%
        \hskip.5\tabcolsep
        \xleaders#3{#2.5\@tempdimb #1{1}#2.5\@tempdimb}%
                #2\z@ plus1fil minus1fil\relax
        \hskip.5\tabcolsep}
\newcommand{\cdashlinelr}[1]{%
  \noalign{\vskip\aboverulesep
           \global\let\@dashdrawstore\adl@draw
           \global\let\adl@draw\adl@drawiv}
  \cdashline{#1}
  \noalign{\global\let\adl@draw\@dashdrawstore
           \vskip\belowrulesep}}
\makeatother

\newcommand{\eg}{e.\,g. }

\newcommand{\cmmnt}[1]{\ignorespaces}

\newcommand{\modelfine}{\textsc{Ft}\xspace}
\newcommand{\modelfinedeb}{\textsc{Ft-Debias}\xspace}
\newcommand{\modelinlp}{\textsc{INLP}\xspace}
\newcommand{\modelinlpnon}{\textsc{INLP-NonLin}\xspace}
\newcommand{\modeladapter}{\textsc{Adp}\xspace}
\newcommand{\modeladapterdeb}{\textsc{Adp-Debias}\xspace}
\newcommand{\modelours}{\textsc{DAM}\xspace}

\title{Parameter-efficient Modularised Bias Mitigation via AdapterFusion}

\author{Deepak Kumar$^1$,~~Oleg Lesota$^1$,~~George Zerveas$^3$,~~Daniel Cohen$^{3,4}$,\\
{\bf Carsten Eickhoff}$^{3,5}$,~~{\bf Markus Schedl}$^{1,2}$,~~{\bf Navid Rekabsaz}$^{1,2}$  \\
  $^1$Johannes Kepler University Linz, Austria\\
  $^2$Linz Institute of Technology, AI Lab\\
  \texttt{\{deepak.kumar,oleg.lesota,markus.schedl,navid.rekabsaz\}@jku.at}\\
  $^3$AI Lab, Brown University, USA \texttt{george\_zerveas@brown.edu}\\
  $^4$Dataminr, USA~~\texttt{daniel.cohen@dataminr.com}\\
  $^5$University of T\"{u}bingen, Germany \texttt{c.eickhoff@acm.org}  
}

\begin{document}
\maketitle
\begin{abstract}
Large pre-trained language models contain societal biases and carry along these biases to downstream tasks. Current in-processing bias mitigation approaches (like adversarial training) impose debiasing by updating a model's parameters, effectively transferring the model to a new, irreversible \emph{debiased} state. In this work, we propose a novel approach to develop stand-alone debiasing functionalities separate from the model, which can be integrated into the model on-demand, while keeping the core model untouched. Drawing from the concept of AdapterFusion in multi-task learning, we introduce \emph{\modelours (Debiasing with Adapter Modules)} -- a debiasing approach to first encapsulate arbitrary bias mitigation functionalities into separate adapters, and then add them to the model on-demand in order to deliver fairness qualities. We conduct a large set of experiments on three classification tasks with gender, race, and age as protected attributes. Our results show that \modelours improves or maintains the effectiveness of bias mitigation, avoids catastrophic forgetting in a multi-attribute scenario, and maintains on-par task performance, while granting parameter-efficiency and easy switching between the original and debiased models. 
\end{abstract}

\section{Introduction}
\label{sec:introduction}


Large pre-trained language models (PLM) and their variations fine-tuned on downstream tasks encode societal biases and stereotypes. A large body of research shows the existence of these biases~\cite{zhao2019gender,sheng2019woman,rekabsaz2021measuring}, and discuss their potential harms~\cite{blodgett2020language} in various tasks such as occupation prediction from biographies~\cite{de2019bias}, information retrieval (IR)~\cite{rekabsaz2021societal,rekabsaz2020neural}, and machine translation~\cite{stanovsky2019evaluating}. A common approach to mitigating these biases is to adjust model parameters according to specific fairness criteria, achieved using optimization methods such as adversarial training~\cite{elazar2018adversarial,barrett2019adversarial,han-etal-2021-diverse,rekabsaz2021societal}.


\begin{figure}[t]
\centering
\includegraphics[width=0.43\textwidth]{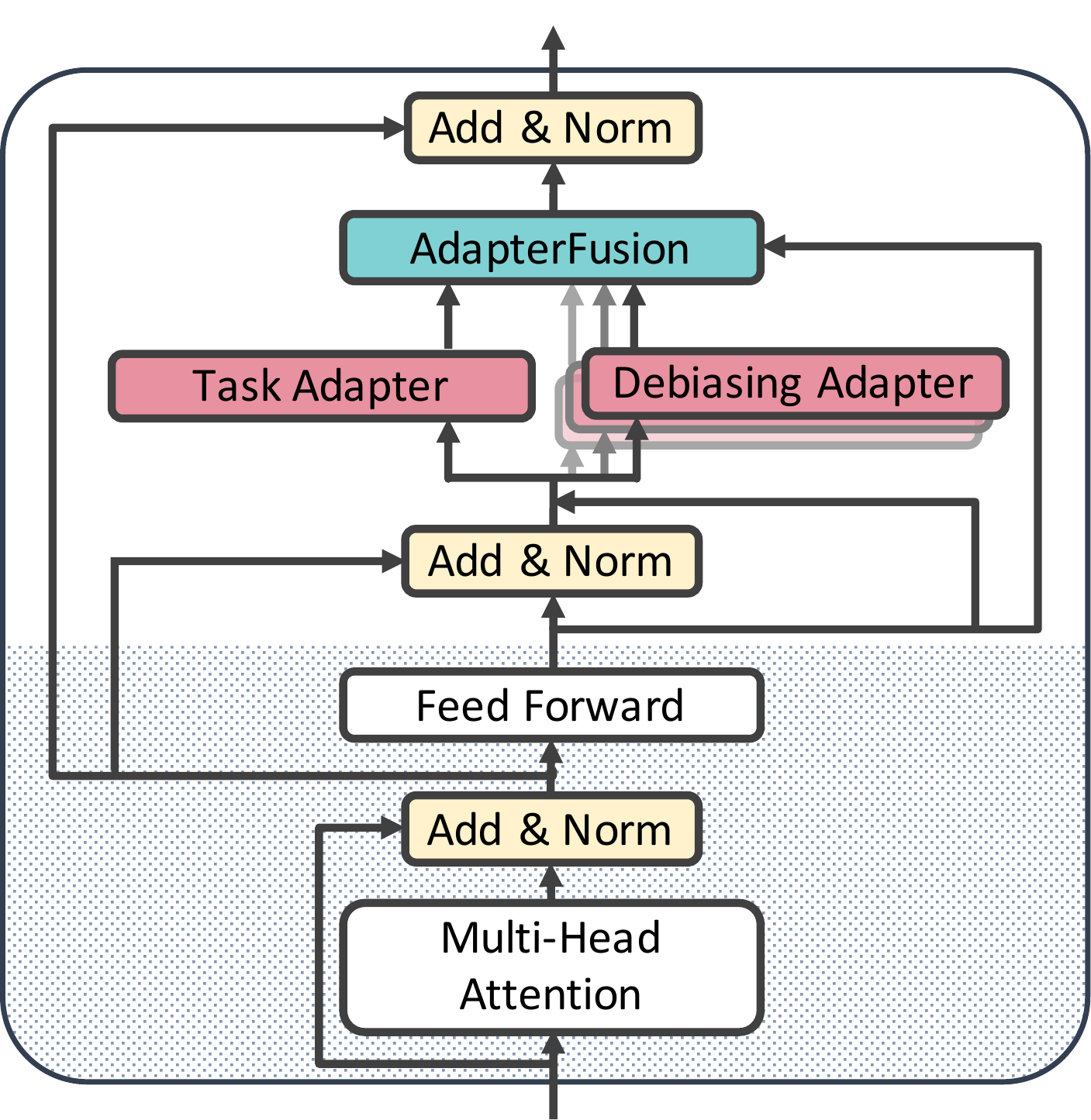}
\caption{A transformer block in \modelours to debias arbitrarily many protected attributes in different modules.}
\label{fig:ecnoder}
\end{figure}
 
The resulting debiased models significantly diminish the effect of protected attributes (\eg gender, race, etc.) on model's predictions. However, depending on the usage context, characteristics of the input or fairness-utility trade-off considerations, system designers or end-users might in practice still prefer to instead use the original (potentially biased) model for processing some requests. As an example, while a fairness-aware document retrieval model should tease out gender attributes when processing bias-sensitive queries~\cite{krieg2022grep} (like \emph{how to become CEO?}), other common queries (like \emph{earliest pregnancy symptoms}) may specifically require gender information to obtain satisfactory retrieval results. Previous studies also show that one may need to occasionally accept slight performance degradation in exchange for realizing higher fairness~\cite{zerveas2022mitigating,biega2018equity,rekabsaz2021societal}. We argue that on-demand control over the strength of bias mitigation is crucial for the broad adoption of bias mitigation models. Using existing approaches, this would require maintaining and deploying multiple large parallel models for every protected attribute, in practice resulting in overly complex and resource-heavy pipelines and increased latency.


To reduce this untenable burden, we introduce \emph{Debiasing with Adapter Modules (\modelours)}. In our approach, bias mitigation functionalities can be trained in fully decoupled modules from the building blocks solving the main task (see Figure~\ref{fig:ecnoder}). These debiasing modules are learned independently, and can be selectively inserted into the model to deliver debiasing qualities. Our method draws from the recent success of Adapter~\cite{rebuffi2017learning,houlsby_2019_param_efficient} and AdapterFusion~\cite{pfeiffer2021adapterfusion} networks, originally introduced in the context of multi-task learning to enable parameter-efficient training and to avoid catastrophic forgetting. Specifically in multi-task learning with AdapterFusion, each task is encoded in specific parameter-efficient adapter networks attached to the transformer blocks~\cite{vaswani_attention_NIPS2017} of a PLM, and the knowledge across the tasks is shared through a FusionAdapter layer with an attention mechanism. \modelours extends this principle idea to bias mitigation by viewing the objective of ``removing each protected attribute'' as a stand-alone task, learned by an independent debiasing adapter module. These debiasing adapters are then integrated into the core model via a fusion layer to mitigate biases. This modular architecture of \modelours enables on-demand debiasing, where by simply removing the debiasing adapters the model would return to its original state. 
The modularity and parameter efficiency of \modelours also enables creating and sharing stand-alone debiasing solutions (in form of debiasing adapters), which can be applied to the underlying models on-demand. So, if one uses task adapter's output in \modelours, they get biased embedding else they can take output of fusion layer for debiased embedding.

We evaluate \modelours in extensive experiments covering various bias mitigation applications in three classification tasks: We study occupation prediction from biographies~\cite{de2019bias}, mention prediction~\cite{Pardo2016OverviewOT}, and hate speech detection~\cite{founta2018large}, involving the protected attributes of gender, age, and dialect-based race. Among these, the mention prediction task provides two protected attributes (gender, age), enabling the study of adapter-based bias mitigation on multiple factors. Our evaluation results show that \modelours (1) consistently improves bias mitigation in terms of lower leakage of protected attribute in comparison with fully fine-tuned models, and (2) better avoids catastrophic forgetting~\cite{parisi2019continual,chen-etal-2020-recall} in a multi-attribute debiasing scenario, while (3) maintaining on-par task performance. Our results show the benefits of \modelours in supporting fairness, and reveal the challenges in the scenarios inherently bounded by trade-offs.



Our contribution is three-fold:
\begin{itemize}
    \item Introducing \modelours, a novel, parameter-efficient approach for on-demand and modularized bias mitigation.
    \item Conducting a large set of experiments on three datasets and three classification tasks, showing the on-par or better performance of \modelours in comparison with strong baselines.
    \item Examining the confounding cross-attribute effects in a multi-attribute dataset and demonstrating the benefits of our modularized bias mitigation approach.
\end{itemize}

The remainder of the paper is organized as follows: Section~\ref{sec:related} reviews the relevant literature in adapter networks and bias mitigation domains. In Section~\ref{sec:method}, we explain the architecture and training strategy of \modelours. Section~\ref{sec:setup} describes the setup of the experiments, whose results are reported and discussed in Section~\ref{sec:results}. 
Finally, Sections~\ref{sec:conclusion} and~\ref{sec:limitations}, respectively, summarize and discuss limitations of our work.
Our code and trained resources are available in \textbf{\url{https://github.com/CPJKU/ModularizedDebiasing}}.

\section{Related Work}
\label{sec:related}

\subsection{Adapter Networks}



Adapter networks have been introduced in the context of multi-task learning~\cite{rebuffi2017learning}. \citet{houlsby_2019_param_efficient}, and later \citet{pmlr-v97-stickland19a} attach the adapter module to the transformer blocks of a PLM, and learn a task by only updating the adapter parameters, while keeping the PLM's parameters unchanged. These studies show that the highly parameter-efficient adapter approach in general performs on par with fine-tuning all parameters of a BERT model~\cite{devlin_2019_bert} for many tasks. 




Other studies investigate various characteristics of adapter-based models such as the parameter efficiency, architectural variations, and transfer learning across tasks. \citet{ruckle2021adapterdrop} show the training efficiency of adapter models in comparison with full model finetuning ($\sim\! 60\%$). \citet{han2021robust} examine the robustness to initialization seeds and training stability of the adapter approach. \citet{DBLP:conf/acl/MahabadiR0H20} propose more parameter-efficient models by sharing adapter parameters across layers, followed by \citet{DBLP:conf/nips/MahabadiHR21}, who introduce even more compact adapter modules. \citet{poth2021pre} investigate the potential knowledge transfer across tasks. Recently, \citet{pfeiffer2021adapterfusion} introduce AdapterFusion, where a fusion layer is defined on top of pre-trained adapter modules, and learns to combine information of adapters via an attention mechanism. This approach, avoiding the common pitfalls of catastrophic forgetting~\cite{parisi2019continual}, enables an effective knowledge transfer across tasks. Our work contributes to this direction by extending the concept of AdapterFusion to the task of bias mitigation via supervised interaction between the downstream task and the protected attribute. 


\subsection{Fairness \& Bias Mitigation in NLP}


The existence of biases and stereotypes in PLMs has been reported and discussed in several studies~\cite{nadeem-etal-2021-stereoset, bhardwaj2021investigating,liang2021towards,kirk2021bias,vig2020investigating}. PLMs may even exacerbate these biases in downstream tasks as shown \eg in the context of IR~\cite{rekabsaz2020neural}. To reduce biases, \textit{in-processing} methods -- the focus of this work -- aim at reducing the (cor)relation between the model's internal representations and the protected attributes~\cite{ganhoer2022mitigating}. Debiasing PLMs has been approached for instance by linearly projecting embeddings to a new space that removes correlations to protected attributes~\cite{kaneko2021debiasing,bolukbasi2016man}. In a similar spirit, \citet{guo2022auto} introduce a distribution alignment loss to force the model's outputs to become independent of the protected attribute. \citet{schick2021self} recently show that the encoded information in models can be exploited to spot the biases and hence to penalize them. 









Adversarial training is a commonly used method to learn representations invariant to a specific factor of variation. \citet{xie2017controllable} and later \citet{madras2018learning} introduce adversarial learning to the context of fair representation learning, where an adversary network learns to predict the protected attribute, and exploits the gradient of this prediction to remove the protected information using a gradient reversal layer~\cite{ganin2015unsupervised}. Several works further investigate the use of adversarial training for removing demographic information from neural/PLM-based text classifiers~\cite{elazar2018adversarial,barrett2019adversarial,wang2021dynamically}. Notably, \citet{han-etal-2021-diverse} show the benefit of having an ensemble of orthogonal adversaries. Beyond classification, \citet{rekabsaz2021societal} show that by applying adversarial training in the context of IR, one can achieve a more balanced retrieval of documents with respect to the presence of protected attributes in their contents.


As alternatives to adversarial debiasing, other works approach bias mitigation by minimizing the approximated upper bound of mutual information between task and protected attribute~\cite{cheng-etal-2020-improving,colombo2021novel}, contrastive learning for fair representation disentanglement~\cite{cheng2020fairfil,zhang2021disentangling}, and introducing list-wise fairness regularization~\cite{zerveas2022mitigating}. While the mentioned methods are applied to whole a model, Iterative Nullspace Projection (INLP)~\cite{ravfogel2020null} achieves debiased representations by finding a linear mapping applied to output embeddings. The linear mapping of the INLP method and similarly the one of \citet{ravfogel2022linear} offer effective bias mitigation methods particularly designed for the models with a linear decoder, but are not necessarily suited for the cases with non-linear decoders (e.g., a language generation decoder, or non-linear attackers). 

Regarding adapter-based debiasing and closely related to our work, \citet{lauscher2021sustainable} recently utilize adapters for debiasing PLMs by training an adapter which removes a protected attribute using counterfactual augmented data. When applied to a downstream task, another adapter is added on top of the debiasing adapter, and trained according to the task's objective. While shown effective in practice, in this stacking architecture the adapters in the higher levels inherently depend on the ones in the lower levels. They can not be learned stand-alone. In contrast, the fusion-based approach of \modelours enables control by learning modular and independent debiasing adapters, which supports flexible plug-in/plug-out on demand.  

\section{Bias Mitigation with \modelours}
\label{sec:method}

We approach the scenario in which the aim is to maximize a task's core objective (such as occupation prediction or hate speech prediction), while reducing the effects of the biases caused by $k$ protected attributes. The crux of the proposed \modelours approach is to first learn a task adapter and $k$ bias removal adapters all independently, and then combine them through a fusion module. Once a model has been trained with \modelours, one can easily switch between the original model by only using the task adapter, or alternatively impose debiasing by ``plugging in'' the learned debiasing adapters as well as the fusion module. In what follows, we explain the architecture of \modelours followed by the optimization procedure.

\subsection{Model Architecture}
\modelours consists of three main parts: task adapter, $k$ debiasing adapters, and fusion. Following \citet{pfeiffer2021adapterfusion}, and as shown in Figure~\ref{fig:ecnoder}, these parts extend the architecture of a transformer block by being added to its last layer.

\paragraph{Adapters} Each adapter is defined as a multilayer perceptron (one hidden layer and $\text{tanh}(x)$ in our experiments), where the hidden layer typically has the same or a smaller dimension than the input. In \modelours, the task adapter and the $k$ debiasing adapters receive the output vector of the preceding transformer components, denoted as $\vu$, and apply the corresponding transformations to output the vectors $\vv_t$, and $\vv_{b_1},...,\vv_{b_k}$, respectively.

\paragraph{Fusion} The fusion module combines the outputs of the task and debiasing adapters through the attention mechanism to produce the final output vector. This module receives $\left[\vv_t,\vv_{b_1},...,\vv_{b_k} \right]$ as keys and values, and the $\vu$ vector as the query of a single-head multiplicative attention network, and calculates the output as the weighted sum of the value vectors. These weights are calculated as attention scores and form a probability distribution over value vectors. Essentially, the fusion module learns to perform a \emph{linear combination} of the embedding containing information for the task with the embeddings of the debiasing adapters, to provide the final, debiased embedding.

\begin{figure}[t]
\centering
\includegraphics[width=0.35\textwidth]{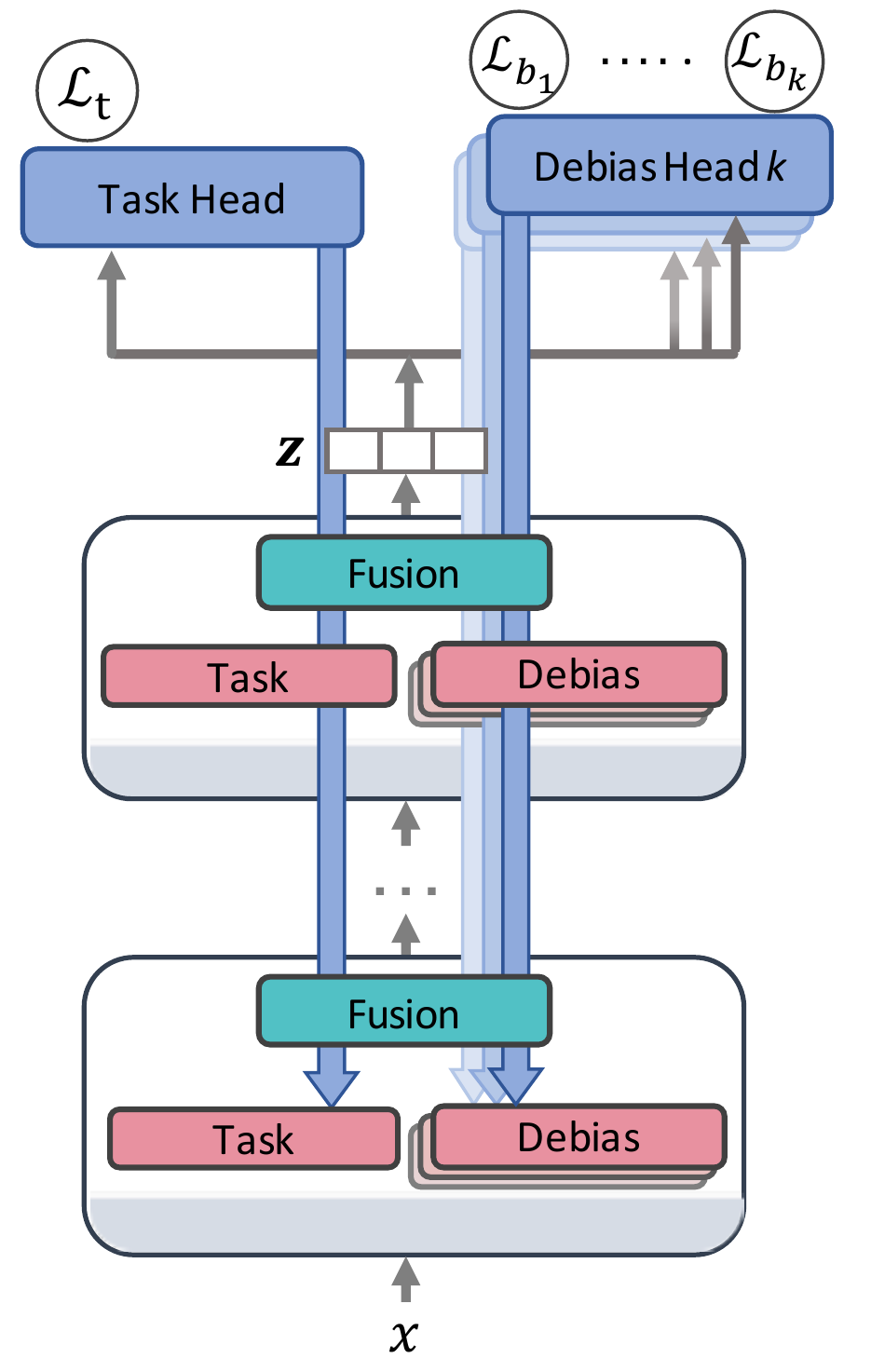}
\caption{Schematic view of applying \modelours to adversarial bias removal.}
\label{fig:architecture}
\end{figure}

\subsection{Learning On-demand Bias Mitigation}
\modelours introduces a generic approach to bias mitigation and representation disentanglement, and can, in principle, be integrated with any bias removal optimization method, provided that the method defines separate losses for the task and each of the protected attributes. Irrespective of the optimization method, the training procedure of \modelours is the following: first, the task adapter is trained according to the task's objective. Then, one debiasing adapter for each protected attribute is trained using only its own debiasing objective (see next paragraph), without involving the task loss. Finally, all adapters' parameters are kept frozen and the fusion layer is trained using a combination of the task objective and debiasing objectives. Throughout training, the parameters of the underlying PLM remain frozen. In what follows, we describe using \modelours together with the adversarial bias mitigation method as depicted in Figure~\ref{fig:architecture}.

\paragraph{Adversarial Training} Adversarial bias mitigation aims to make the output embedding of a model invariant to the protected attribute, by removing information that allows predicting protected attributes from the latent representation. In this sense, adversarial training follows a \emph{fairness through blindness}~\cite{barocas-hardt-narayanan} approach, where the system is made agnostic of the variations in underlying protected attributes. 

To apply adversarial debiasing with \modelours (see Figure~\ref{fig:architecture}) in the context of text classification, the model first encodes the input sequence $x$ in the latent vector $\vz$, based on which the corresponding class is predicted using a task classification head. The loss of this prediction, denoted as $\Ls_{t}$, is defined as cross entropy, and its gradient is used to update the parameters of the task adapter. Similarly, a dedicated classification head is defined for each protected attribute $b_i$, which also receives $\vz$ as input, predicts the corresponding protected attribute, and calculates the cross entropy loss function $\Ls_{b_i}$. A common approach to removing the information of $b_i$ encoded in $\vz$ is gradient resversal layer (GRL)~\cite{ganin2015unsupervised} added before the debiasing head. GRL multiplies the gradient of $\Ls_{b_i}$ with a factor of $-\gamma_{i}$, and thereby simplifies the learning process to a standard gradient-based optimization. In \modelours, this reversed gradient of $\Ls_{b_i}$ is used to learn the parameters of the debiasing adapter corresponding to the protected attribute $b_i$. Once adapters are trained, the fusion layer is learned jointly according to all task and debiasing objectives, as formulated below:
\begin{equation}
    \Ls_{\text{Fusion}}^{\text{Adv}} = \Ls_{t}+\sum_{i=1}^{k}{\Ls_{b_i}}
\end{equation}
Note that, while no weights are used to directly scale the individual loss functions, the effects of bias mitigation losses on model parameters are adjusted via their corresponding $\gamma_i$ hyperparameters.

\section{Experiment Setup}
\label{sec:setup}

We evaluate \modelours on three classification tasks and compare the results with a set of strong baselines.  

\paragraph{Tasks and Data} We conduct our experiments on three classification tasks. The first task is \textbf{biography classification} using the BIOS~\cite{de2019bias} dataset. The objective of the task is to predict a person's job given their biography, where the name and any indication of the person's gender (such as pronouns) in the biography is omitted. The protected attribute is the person's gender. The BIOS dataset contains around 430K data points with 28 occupations as the core task's classes, and two protected attribute classes (female/male). 

The second task is \textbf{hate speech detection}, a sensitive task, as its use in any automated system with biases towards any societal group can be extremely delicate (\eg in automatic social media moderation). \citet{xia-etal-2020-demoting} show the existence of a strong correlation between dialect-based racial bias and hate speech detection. In our experiments, we use FDCL18~\cite{founta2018large}, a dataset of tweets annotated with four hate speech labels: \emph{normal}, \emph{spam}, \emph{abusive}, and \emph{hateful}. Following previous works~\citep{sap-etal-2019-risk,ravfogel2020null, zhang2021disentangling}, we assign artificial race labels of \emph{African American}, and \emph{White American} to FDCL18 (more details in Appendix~\ref{sec:appendix:setting:dataset}). In this setting, the main task is hate speech classification, and the protected attribute is the dialect-based racial identity of the tweet author. Our final dataset contains approximately 62K data points.

\begin{table}[t]
\centering
\scalebox{0.8}{
\begin{tabular}{ l l r r }
\toprule
\multicolumn{2}{l}{Model} & \multicolumn{1}{c}{\# Params}  \\\midrule
\multirow{5}{*}{BERT-Mini} & \modelfine & $11,237,380$   \\
& \modelfinedeb & $11,568,910$ \\
& \modeladapter & $330,500$  \\
& \modeladapterdeb & $432,206$ \\
& \modelours & $1,186,830$ \\\midrule
\multirow{5}{*}{BERT-Base} & \modelfine & $110,075,908$   \\
& \modelfinedeb & $113,036,558$  \\
& \modeladapter & $14,767,876$  \\
& \modeladapterdeb & $4,448,846$ \\
& \modelours & $24,806,414$ \\
\bottomrule
\end{tabular}
}
\caption{Number of trainable parameters. The adapter-based models contain significantly fewer trainable parameters in comparison with fully fine-tuned models.}
\label{tbl:parameter}
\vspace{-0.3cm}
\end{table}

While the first two tasks contain one protected attribute, in our third classification task, we aim to explore the effect of more than one biases on our models. We therefore opt for a subset of the PAN16 dataset~\cite{Pardo2016OverviewOT}. 
This dataset has been designed for the task of \textbf{mention detection}, namely to predict whether a tweet includes the mention of other users, and the protected attributes are gender and age of the tweet's author. The dataset contains approximately 200K data points with binary task classes (\emph{mention}, \emph{no mention}), as well as two gender labels and five age groups. Further details on the three datasets are provided in Appendix~\ref{sec:appendix:setting:dataset}.

\begin{table*}[t]
\centering
\begin{tabular}{ l l ll || ll}
\toprule

&& \multicolumn{2}{c}{\textbf{BIOS (gender)}} & \multicolumn{2}{c}{\textbf{FDCL18 (race)}} \\

 \multirow{1}{*}{Model} && \multirow{1}{*}{\textbf{Task}$\uparrow$} & \multicolumn{1}{c}{\textbf{Attacker}$\downarrow$} & \multirow{1}{*}{\textbf{Task}$\uparrow$} & \multicolumn{1}{c}{\textbf{Attacker}$\downarrow$}
\\\midrule

\multirow{7}{*}{BERT-Mini} & \modelfine & $82.81^{\clubsuit}_{0.1}$ & $65.10_{0.3}$& $81.83^{\clubsuit}_{0.4}$& $92.72_{1.1}$  \\
& \modelfinedeb & $81.52_{0.2}$	&	$57.34^{\clubsuit}_{0.8}$ &	$79.15_{0.8}$	&	$76.14_{1.8}$ \\
& \modelinlp & $71.08_{0.2}$  & \cmmnt{$60.11_{0.8}$} $59.83_{0.9}$ & $73.60_{1.6}$   \cmmnt{$92.13_{1.4}$} & $71.75^{\clubsuit}_{0.4}$  \\
& \modelinlpnon & $54.17_{2.1}$  \cmmnt{$60.23_{1.2}$} & $59.90_{1.3}$ & $51.49_{0.2}$  \cmmnt{$92.19_{0.0}$} & $92.19_{0.0}$\\
\cdashlinelr{2-6}
& \modeladapter &  $81.82_{0.1}$	&	$66.59_{0.2}$ &	$81.16_ {0.2}$	&	$93.69_{0.5}$   \\
& \modeladapterdeb & $81.71_{0.1}$	&	$62.97_{0.3}$	&	$\textbf{81.46}_{0.1}$	&	$81.96_{0.9}$ \\
& \modelours & $\textbf{81.97}_{0.1}$	&	$\textbf{57.78}_{0.3}$	&	$81.06_{0.2}$	&	$\textbf{77.52}_{1.4}$\\
\midrule

\multirow{7}{*}{BERT-Base} & \modelfine & $85.07^{\clubsuit}_{0.1}$	&	$67.15_{0.4}$	&	$84.83^{\clubsuit}_{0.5}$	&	$95.87_{0.6}$   \\
& \modelfinedeb & $84.21_{0.2}$	&	$58.35_{1.1}$	&	$84.15_{0.8}$	&	$72.95^{\clubsuit}_{0.8}$  \\
& \modelinlp & $66.25_{0.4}$  \cmmnt{$52.42^{\clubsuit}_{0.8}$} & $50.51^{\clubsuit}_{0.3}$ & $71.63_{2.2}$ & $69.87_{0.2}$\\
& \modelinlpnon &  $24.44_{3.3}$ \cmmnt{ $55.41_{1.1}$} & $54.81_{0.9}$ & $42.23_{0.4}$ &  $91.67_{0.3}$\\
\cdashlinelr{2-6}
& \modeladapter &  $84.84_{0.1}$	&	$67.80_{0.2}$	&	$82.25_{0.4}$	&	$95.89_{0.4}$  \\
& \modeladapterdeb & $84.40_{0.1}$	&	$60.94_{0.4}$	&	$83.30_{0.3}$	&	$77.97_{2.0}$\\
& \modelours & $\textbf{84.86}_{0.1}$	&	$\textbf{58.93}_{0.2}$	&	$\textbf{84.76}_{0.4}$	&	$\textbf{73.54}_{0.4}$\\

\bottomrule
\end{tabular}
\caption{Results on BIOS and FDCL18 datasets for the task classifiers (accuracy) and protected attributes' attackers (balanced accuracy), reported as the mean and standard deviation (in $_{\mathrm{subscript}}$) over three runs. Arrows indicate the direction of better results. On each evaluation metric and each PLM, the best results among adapter-based models are shown in \textbf{bold}, and over all models by symbol $\clubsuit$.}
\label{tbl:results:bios_hatespeech}
\end{table*}

\begin{table*}[t!]
\centering
\begin{tabular}{l l r r r  }
\toprule
Model & & {\textbf{Task}$\uparrow$} & \multicolumn{2}{c}{\textbf{Attackers}$\downarrow$} \\
& & \multicolumn{1}{c}{} & \multicolumn{1}{c}{\textbf{Gender}}  & \multicolumn{1}{c}{\textbf{Age}}
 \\\midrule
 
\multirow{15}{*}{BERT-Mini} & \modelfine & $	90.43_{0.5}$	&	$	64.36_{0.6}$	&	$	30.77_{0.9}$\\
& $\modelfinedeb_{G}$ & $91.89^{\clubsuit}_{0.3}$	&	$	54.60_{0.4}$	&	$	25.50_{0.6}$\\
& $\modelfinedeb_{A}$ & $	91.06_{0.1}$	&	$	57.27_{1.3}$	&	$	22.29_{0.6}$ \\
& $\modelfinedeb_{G+A}$ & $	90.74_{0.5}$	&	$	59.13_{0.5}$	&	$	24.93_{0.3}$ \\
& $\modelinlp_{G}$ & $69.26_{0.1}$ & $60.06_{0.3}$ & $29.21_{0.8}$ \\
& $\modelinlpnon_{G}$ & $67.00_{0.1}$ &  $60.35_{0.2}$& $30.03_{0.1}$ \\
& $\modelinlp_{A}$ & $66.72_{1.8}$ &  $60.62_{0.1}$ &  $25.63_{0.2}$ \\
& $\modelinlpnon_{A}$ & $62.21_{6.6}$ & $60.39_{0.3}$ & $25.51_{0.3}$ \\

\cdashlinelr{2-5}
& \modeladapter & $	89.97_{0.2}$	&	$	68.68_{0.3}$	&	$	35.97_{6.8}$\\
& $\modeladapterdeb_{G}$ & $	88.53_{0.2}$	&	$	56.92_{1.4}$	&	$	29.58_{1.1}$ \\
& $\modeladapterdeb_{A}$ & $	89.90_{0.1}$	&	$	62.32_{0.6}$	&	$	24.75_{0.2}$ \\
& $\modeladapterdeb_{G+A}$ & $	88.87_{0.1}$	&	$	56.99_{0.4}$	&	$	26.72_{0.1}$ \\
& $\modelours_{G}$ & $	\textbf{90.52}_{0.2}$	&	$	54.35_{0.6}$	&	$	26.09_{0.6}$\\
& $\modelours_{A}$ & $	90.31_{0.2}$	&	$	61.78_{0.2}$	&	$	21.97_{0.1}$\\
& $\modelours_{G+A}$ & $90.19_{0.2}$	&	$	\textbf{53.28}^{\clubsuit}_{0.6}$	&	$	\textbf{21.91}^{\clubsuit}_{0.3}$ \\
\midrule

\multirow{15}{*}{BERT-Base} & \modelfine & $	92.17_{0.5}$	&	$	71.10_{1.8}$	&	$	42.29_{7.4}$\\
& $\modelfinedeb_{G}$ & $	92.80_{0.3}$	&	$	53.80_{0.4}$	&	$	26.92_{2.3}$\\
& $\modelfinedeb_{A}$ & $	93.79^{\clubsuit}_{0.1}$	&	$	56.26_{0.7}$	&	$	21.89_{0.5}$ \\
& $\modelfinedeb_{G+A}$ & $	92.39_{0.4}$	&	$	51.96_{0.1}$	&	$	27.43_{5.4}$ \\
& $\modelinlp_{G}$ & $69.37_{0.1}$ & $54.40_{0.1}$ & $25.53_{0.2}$ \\
& $\modelinlpnon_{G}$ & $65.42_{0.1}$ &   $54.11_{0.1}$ & $26.98_{0.2}$ \\
& $\modelinlp_{A}$ & $49.82_{0.8}$ &  $54.61_{0.6}$ & $27.55_{0.2}$\\
& $\modelinlpnon_{A}$ & $50.18_{5.3}$ & $54.44_{0.4}$ & $33.66_{0.0}$ \\
\cdashlinelr{2-5}
& \modeladapter & $	92.34_{0.4}$	&	$	70.00_{0.3}$	&	$	42.93_{1.9}$\\
& $\modeladapterdeb_{G}$ & $	\textbf{93.32}_{0.2}$	&	$	54.17_{0.2}$	&	$	29.67_{0.3}$ \\
& $\modeladapterdeb_{A}$ & $	88.88_{0.2}$	&	$	54.48_{0.6}$	&	$	21.96_{0.6}$ \\
& $\modeladapterdeb_{G+A}$ & $	89.91_{1.8}$	&	$	52.92_{0.8}$	&	$	21.79_{0.2}$\\
& $\modelours_{G}$ & $	92.32_{0.3}$	&	$	52.75_{0.6}$	&	$	28.12_{0.3}$\\
& $\modelours_{A}$ & $	92.06_{0.4}$	&	$	59.18_{0.4}$	&	$	\textbf{21.24}^{\clubsuit}_{0.2}$\\
& $\modelours_{G+A}$ & $	92.44_{0.4}$	&	$	\textbf{51.58}^{\clubsuit}_{0.5}$	&	$	21.95_{0.5}$ \\

\bottomrule
\end{tabular}

\caption{Results of the PAN16 mention detection task with gender ($G$) and age ($A$) as protected attributes on the task classifiers (accuracy) and protected attributes' attackers (balanced accuracy). The mean and standard deviation (in $_{\mathrm{subscript}}$) over three runs are reported. On each evaluation metric and each PLM, the best results among adapter-based models are shown in \textbf{bold}, and over all models by symbol $\clubsuit$.}
\label{tbl:results:pan16}
\end{table*}


\vspace{-2mm}
\paragraph{Models and Baselines}
We conduct the experiments on the following models: \textbf{\modelfine}: fully fine-tuning the PLM on the task. \makebox{\textbf{{\modelfinedeb}}}: fully fine-tuning the PLM on the task as well as on the corresponding adversarial debiasing objective(s). \makebox{\textbf{\modelinlp}}: using the implementation and suggested hyperparameter setting by~\citet{ravfogel2020null}, in particular with a linear task classifier (Logistic Regressor). \makebox{\textbf{\modelinlpnon}}: the \modelinlp model with the same setting, but instead of Logistic Regressor, \modelinlpnon uses a non-linear task classifier: a two-layer feed-forward layer with a tanh activation. This setting follows the same configuration of all other baselines/models with respect to the task classifier. \textbf{\modeladapter}: adding one adapter to the PLM and training it only through the task objective. \makebox{\textbf{\modeladapterdeb}}: adding one adapter to the PLM and training it through the task objective as well as the adversarial debiasing objective(s). Finally, our proposed \textbf{\modelours} model, which consists of one task adapter and separate debiasing adapters for each protected attribute, all combined with a fusion module. \modelours uses the same task adapter trained by \modeladapter and further learns the debiasing adapters and fusion layer. We should note that, since through adapter-based training the task performance of \modeladapter might be different from the one of \modelfine, a fair comparison of the core task performance should be between the results of \modeladapter and its respective \modelours model. We emphasize that we are especially interested in adapters because they preserve the original model and can be disengaged at will, to trade off debiasing for utility/performance.

As the PLM encoder for all models, we conduct our experiments using two versions of BERT~\cite{devlin_2019_bert} with different sizes, namely BERT-Mini~\cite{turc2019wellread} and BERT-Base in order to provide a more comprehensive picture regarding the effect of encoder size and number of involved parameters. The number of trainable parameters of the models is reported in Table~\ref{tbl:parameter}. The complete details of our hyperparameters setting and training procedure are explained in Appendix~\ref{sec:appendix:setting:hyperparam} and Appendix~\ref{sec:appendix:setting:training}, respectively.

\vspace{-2mm}
\paragraph{Evaluation and Attackers}
To evaluate the performance of the classifiers on the core task, we use the accuracy metric. To evaluate bias mitigation, following previous works~\cite{elazar2018adversarial,barrett2019adversarial}, we report the leakage of a protected attribute in terms of accuracy and balanced accuracy of an \emph{attacker}. To train an attacker, we freeze the model's parameters, and train a new classification head (two-layer feed-forward layer with a $\text{tanh}(x)$ activation) to predict the protected attribute from the $\vz$ vector. We train an ensemble of 5 attackers with the same configuration and training procedure, and report the results of the best performing one. We report the performance of the attacker in terms of accuracy and also balanced/macro accuracy (average of per-class accuracy scores). Balanced accuracy has the benefit of better reflecting the performance of the methods when considering sub- and minority groups. We believe this is particularly important in this setting, since the datasets are unbalanced over both task and protected labels (see Appendix~\ref{sec:appendix:setting:dataset}). To account for possible variabilities in training, we repeat every experiment three times and report the mean as well as standard deviation of the achieved results.

\section{Results and Discussion}
\label{sec:results}

In this section, we first discuss the results on the datasets with one protected attribute (BIOS and FDCL18), followed by investigating the effect of \modelours on PAN16 with two protected attributes.

\paragraph{Single-attribute bias mitigation}

The results on BIOS and FDCL18 datasets are reported in Table~\ref{tbl:results:bios_hatespeech}. We start with the results of BERT-Mini on both datasets. As shown, consistent with previous studies the task performance of \modeladapter is slightly worse then the full model fine-tuning \modelfine. In both \modelfine and \modeladapter, we observe that the adversarial debiasing (\modelfinedeb and \modeladapterdeb respectively) significantly reduces leakage in terms of Acc and BAcc of attackers, while task performance remains on par with the corresponding baseline models. Overall, adversarial methods outperform respective \modelinlp models, particularly in respect to maintaining task performance during bias mitigation.

Compared to \modeladapter and \modeladapterdeb, we observe that \modelours improves \modeladapterdeb in terms of bias mitigation, while maintaining or even slightly improving performance on the task. The difference between \modelours and \modeladapterdeb indicates the benefit of learning separate adapters for performing the task and bias mitigation, and then combining them. Also comparing \modelours with \modelfinedeb, overall \modelours performs on par in terms of task performance and leakage, while \modelours provides the additional benefits of selectively engaging bias mitigation and significantly fewer trainable parameters.

Comparing the results between the core PLMs, we see that the above observations for BERT-Mini also hold true for BERT-Base on the BIOS dataset and the FDCL18 dataset.


\paragraph{Multi/Two-attribute bias mitigation}
Table~\ref{tbl:results:pan16} reports the results on PAN16 with two protected attributes, gender ($G$) and age ($A$). To observe the effect of single- versus multiple-attribute debiasing, we train the debiasing models (\modelfinedeb, \modeladapterdeb, and \modelours) in three variations denoted with subscripts $G$, $A$, and $G\!\!+\!\!A$, indicating the debiasing optimization on only gender, only age, and both gender and age, respectively.\footnote{We run \modelinlp and \modelinlpnon on the two variations of $G$ and $A$, but we are not aware of a principled way for this method to simultaneously debias more than one protected attribute.} 

Observing similar patterns to the previous experiments, $\modelours_{G}$, $\modelours_{A}$, and $\modelours_{G+A}$ provide a significant improvement in bias mitigation (lower leakage) in comparison with their respective debiasing baseline variations. Also similar to single-attribute datasets, \modeladapter performs slightly worse than \modelfine in respect to task performance, but \modelours models remain on par with \modeladapter or slightly improve it. We observe these patterns consistently on both BERT-Mini and BERT-Base. Overall, \modelours provides equal or better bias mitigation in comparison with \modelfinedeb and \modelinlp, while offering on-demand debiasing and parameter efficiency.

Particularly with respect to multi-attribute bias mitigation, the results show the benefits of $\modelours_{A+G}$ in separately learning multiple debiasing adapters and combining them, when compared with $\modelfinedeb_{G+A}$, which simultaneously learns both debiasing factors. In particular, looking at the attacker's BAcc, while $\modelfinedeb_{G+A}$ is not able to debias the age attribute (in comparison to \modelfine), our $\modelours_{G+A}$ provides strongly improved bias mitigation for age (and also gender), with similar or better results to only optimizing for gender or age ($\modelours_{G}$ and $\modelours_{A}$, respectively). This highlights the effectiveness of \modelours in preventing catastrophic forgetting of debiasing functionalities in multi-attribute scenarios.\footnote{Our experiments using \modelours also indicate a possible correlation between Gender and Age, as debiasing Gender ($\modelours_{G}$) leads to slight reduction in Age leakage, while similarly, debiasing Age ($\modelours_{A}$) results in a slight removal of gender information.} We further provide an investigation of the attention distributions of \modelours's fusion module in Appendix~\ref{sec:appendix:fusion}.





\section{Conclusion}
\label{sec:conclusion}
We propose a novel bias mitigation approach which enables flexible switching between the original and debiased state of a model. Our proposed \modelours method extends the idea of multi-task learning using AdapterFusion to bias mitigation, by first learning the main task and the debiasing objectives as separate adapters, and then leveraging the attention-based fusion approach to merge these adapters and deliver debiased results. Our experiments on three classification tasks show that, beside flexible switching, \modelours improves bias mitigation, provides on par classification performance to vanilla adversarial debiasing, and addresses the issue of catastrophic forgetting in multi-attribute bias mitigation. 

\section{Limitations}
\label{sec:limitations}

In the experiments, gender is considered as a binary construct due to practical constraints. In particular, in BIOS and PAN16 the gender is provided only in the form of male and female. We are however fully aware that a gender binary model is not representative of all individuals; yet working with in-the-wild data (as in our case) entails an unavoidable caveat, derived from the still predominant belief that human beings can be sorted into two discrete categories~\cite{hyde2019future}. However, the proposed method can still be defined for generic non-binary settings and can be applied to any sensitive attribute with more than two categories, as exemplified by our consideration of age classes.


We should also highlight two limitations of this study in respect to the model and optimization. First, adversarial approaches in general (including our proposed approach) aims to reduce the correlations in the model to the protected attribute based on the observed data. This approach, like other data-oriented bias mitigation methods, might lack effective generalization, particularly when the model is evaluated on other domains or out-of-distribution data. The second limitation regarding the discussed bias mitigation methods based on adversarial training (also involving our proposed approach) is that, while the method aims to make the model prediction agnostic to protected attributes, it does not directly account for a balanced treatment of subpopulations in regards to the utility metrics and quality of the received service. 

\section{Acknowledgment}
This work received financial support by the Austrian Science Fund (FWF): P33526 and DFH-23; and by the State of Upper Austria and the Federal Ministry of Education, Science, and Research, through grants LIT-2020-9-SEE-113 and LIT-2021-YOU-215.

\bibliography{references}
\bibliographystyle{acl_natbib}

\appendix
\section{Experiment Settings -- Additional Details}
\label{sec:appendix:setting}

\subsection{Datasets}
\label{sec:appendix:setting:dataset}
In FDCL18 dataset, we use the TwitterAAE model~\cite{blodgett-etal-2016-demographic} to assign racial dialect classes. The TwiiterAAE model predicts four racial classes, \emph{African American}, \emph{White American}, \emph{Hispanic}, and \emph{Others}. We labeled a tweet as \emph{African American} or \emph{White American} if the prediction score was greater than $0.5$. For PAN16 dataset, following \cite{sap-etal-2019-risk} we balanced the task labels and sampled 200K data. The age groups of this dataset are 18-24, 25-34, 35-49, 50-64, and 65+. The proportions of data items regarding the labels of the task and protected attributes in the three dataset are as follows:
\begin{itemize}
    \item BIOS dataset: Task (dentist: 0.03, professor: 0.27, teacher: 0.04, psychologist: 0.04, nurse: 0.07, poet: 0.01, photographer: 0.06, journalist: 0.04, filmmaker: 0.02, physician: 0.08, composer: 0.2, attorney: 0.08, model: 0.03, painter: 0.02, accountant: 0.01, pastor: 0.01, comedian: 0.01, surgeon: 0.05, architect: 0.03, paralegal: 0.01, dj: 0.01, chiropractor: 0.01, software engineer: 0.02, dietitian: 0.02, rapper: 0.01, personal trainer: 0.003, yoga teacher: 0.01, interior designer: 0.01); Gender (Female: 0.5, Male: 0.5)
    \item FDCL18 dataset: Task (normal: 0.73, spam: 0.12, abusive: 0.12, hateful: 0.04); Race (White: 0.5, AA: 0.5)
    \item PAN16 dataset: Task (Mention: 0.5, Not Mention: 0.5); Gender (male: 0.54, female: 0.46); Age ( 3: 0.18, 1: 0.34, 2: 0.40, 0: 0.07, 4: 0.01)
\end{itemize}


\subsection{Hyperparameter setting}
\label{sec:appendix:setting:hyperparam}
Across experiments, we keep specific hyperparameters consistent. Batch size is 64, learning rate is 2e-5 (except for training task and debiasing adapter as explained below), training epochs is 20, dropout probability is 0.3, and adversarial debiasing coefficient is 1 for all models (when applicable). 

For task adapter and debiasing adapters, we tune the learning rate and the hidden layer dimension of adapter. We conduct brute search over the learning rate values of [1e-5, 1e-4,1e-3,1e-2], and hidden layer dimension with a division factor of [1,1/2,1/4,1/8,1/16].
For \modelinlp we use 10 iterations and 10 classifiers to learn null space while for \modelinlpnon we same setting (300 iterations and 300 classifiers) as in \cite{ravfogel2020null}.

\subsection{Training procedure}
\label{sec:appendix:setting:training}
We randomly split the dataset into train, validation, and test set with the proportions 63:12:15 for BIOS, 63:12:15 for FDCL18, and 80:5:15 on PAN16. We use the validation set for hyperparameter tuning, and the best result on the validation set is evaluated on test set for the final results. The validation and test sets in all datasets follow the same distribution as the whole dataset. To address the unbalancedness of the dataset and the potential problems in adversarial learning, we apply upsampling only on the \emph{training sets} of BIOS and FDCL18 datasets, to balance the protected attribute labels within each task label. For instance, genders are balanced in the dentist class by repeating the data items of the minority subgroup.

\section{Fusion Attention Analysis}
\label{sec:appendix:fusion}
\begin{figure*}[t]
\centering
\subfloat[BIOS]{\includegraphics[width=0.45\textwidth]{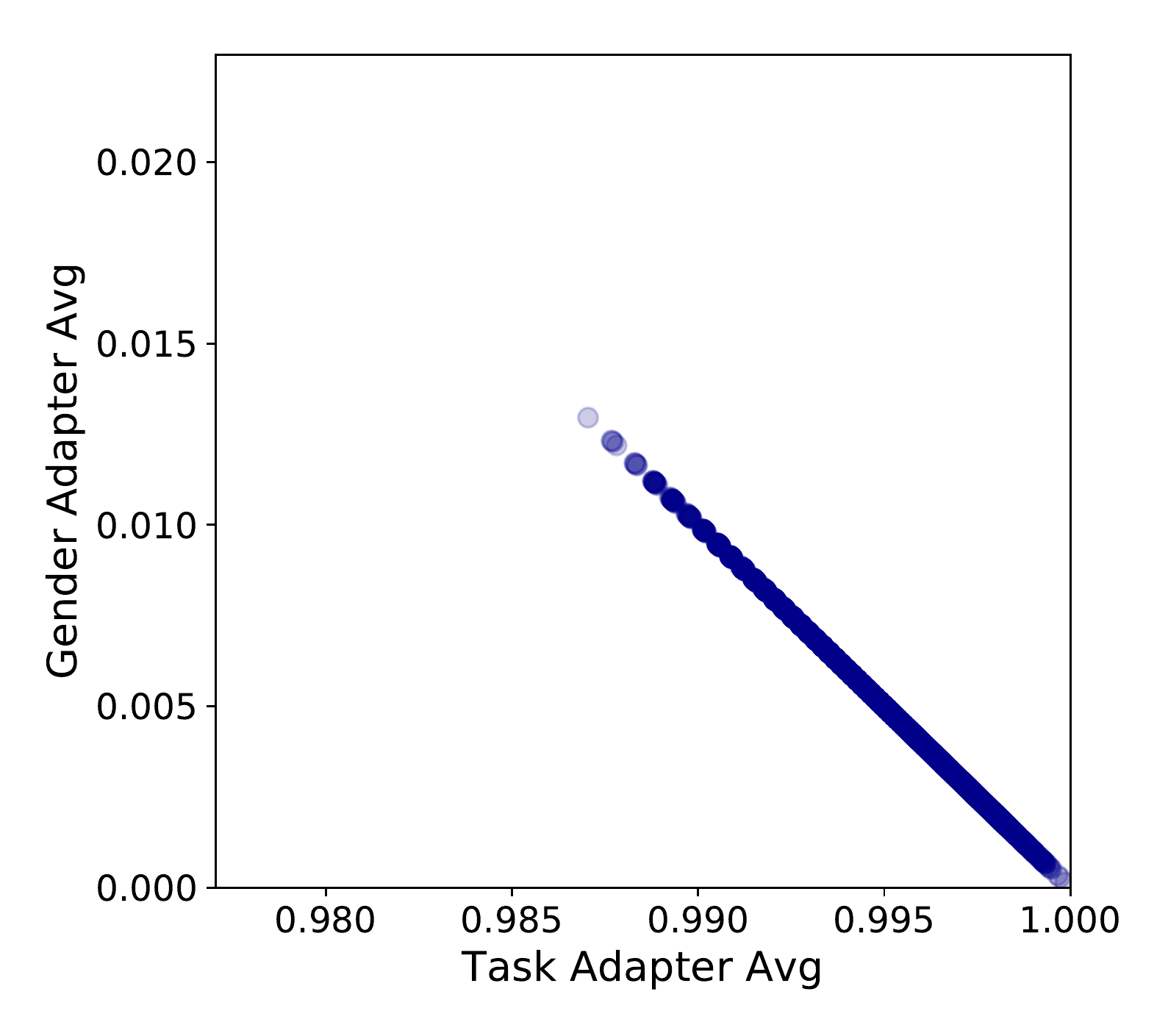}\label{fig:appendix:bio}}
\hspace{10mm}
\subfloat[FDCL18]{\includegraphics[width=0.44\textwidth]{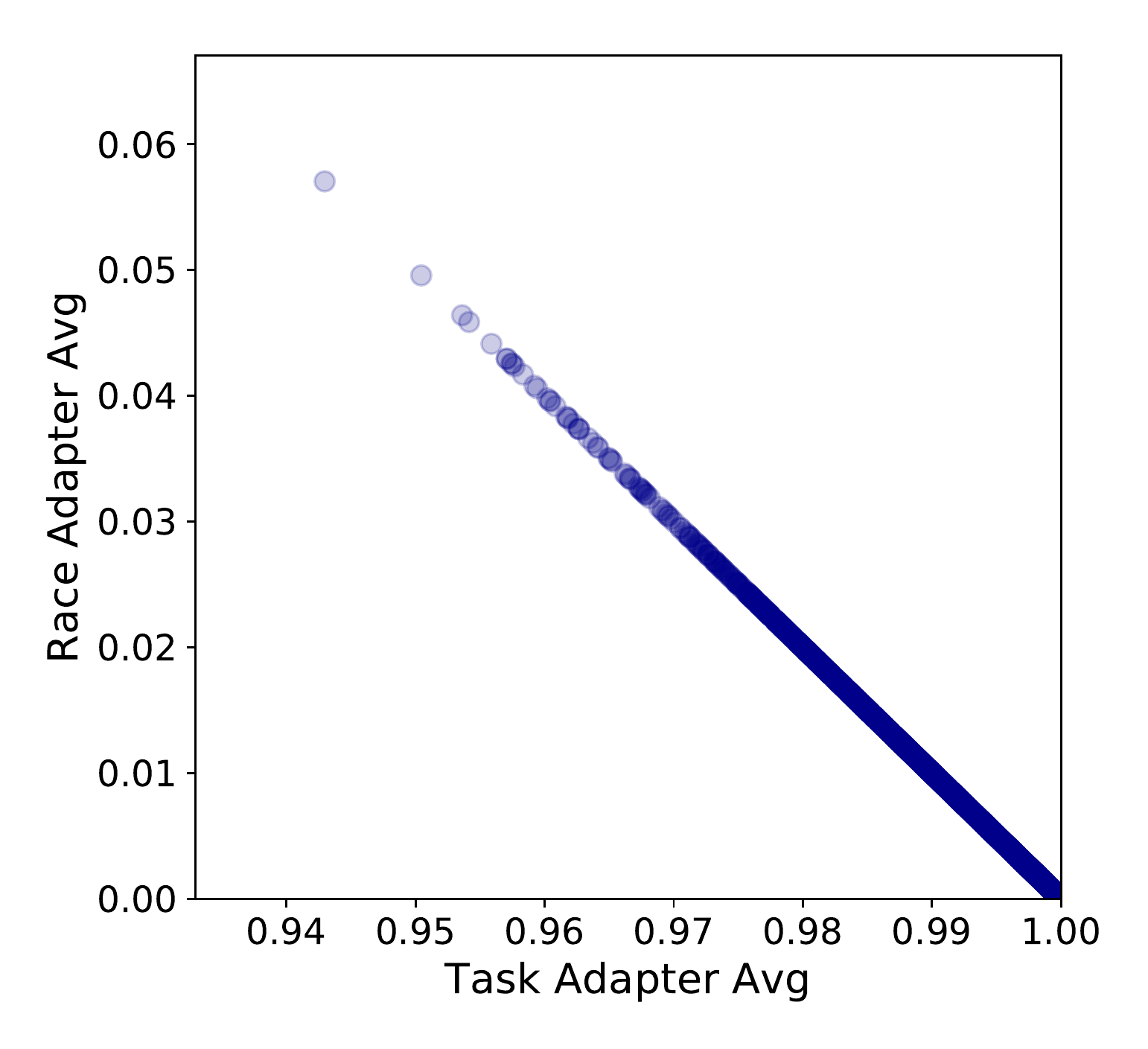}\label{fig:appendix:race}}
\centering
\caption{Attention distribution of a set of sampled data points in the fusion module of \modelours.}
\label{fig:appendix:bio_race}
\end{figure*}

\begin{figure*}[t]
\centering
\includegraphics[width=0.4\textwidth]{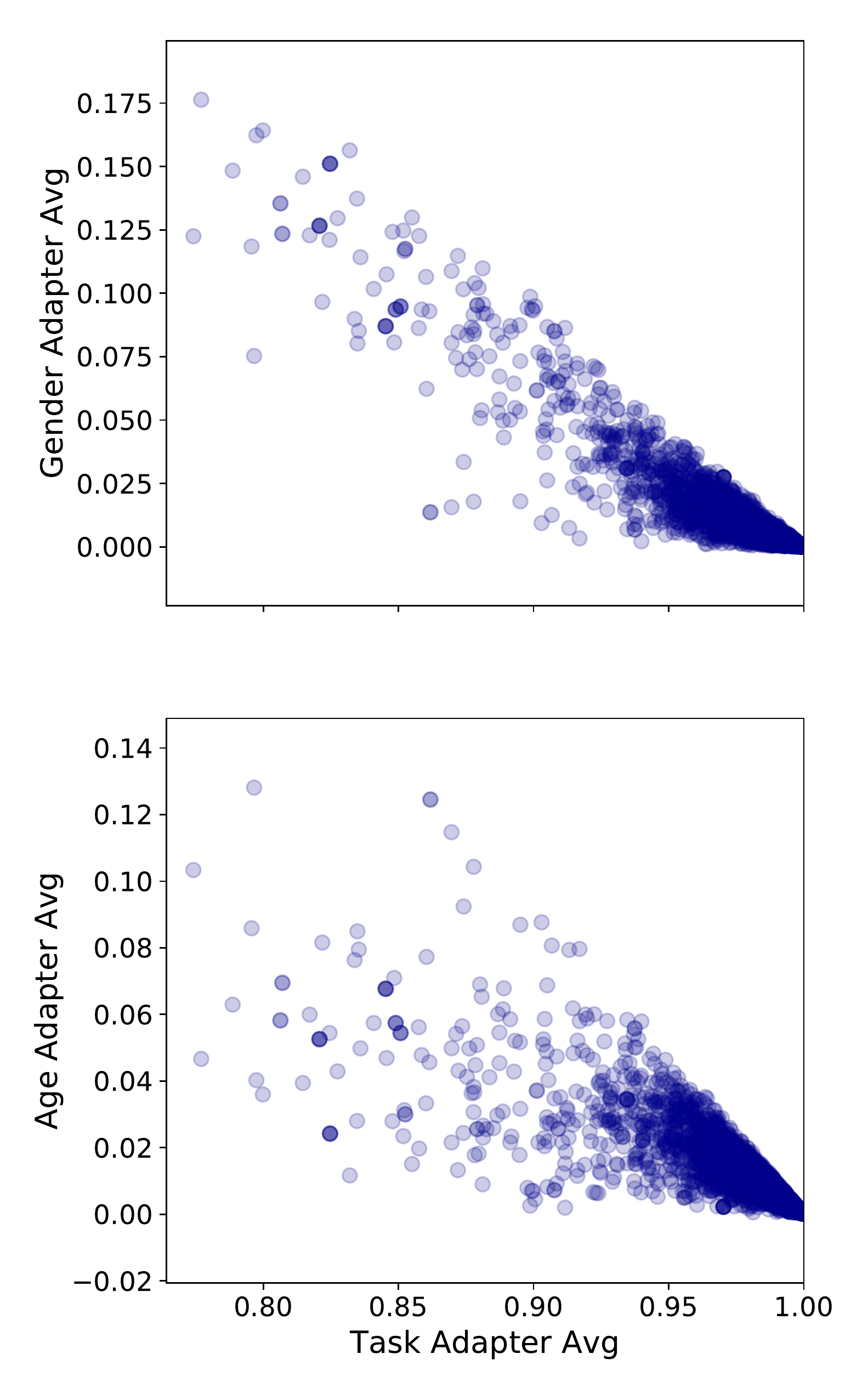}
\caption{Attention distribution of a set of sampled data points in the fusion module of \modelours in PAN16 dataset.}
\label{fig:appendix:mention}
\end{figure*}

We investigate the attention distribution of the fusion network and observe the weights it gives to the adapters. Figures~\ref{fig:appendix:bio}, \ref{fig:appendix:race}, and \ref{fig:appendix:mention} depict the attention scores of each adapter averaged over all fusion layer in \modelours for BIOS, and FDCL18, and PAN16 datasets, respectively. To avoid confusion in visualization, we only used 4\% of data points randomly sampled from test set. As shown, the task adapter has weights close to 1, while debiasing adapters are assigned attention scores slightly higher than 0. The top three outliers in BIOS with the highest attentions on the debiasing adapter are reported in Table~\ref{tbl:appendix:datapoints}.

\begin{table}[h]
\centering
\begin{tabular}{ l L{6cm} }
\toprule
\multirow{1}{*}{} & \multicolumn{1}{c}{\textbf{Text}} \\\midrule

1 & passionately promotes healthy dietary and lifestyle choices to prevent disease and achieve optimal health \\
2 & practices include clayton heights family dental street panorama family dental avenue and surrey family dental avenue \\
3 & primary research interests include collective security and global health in an international relations and international political economy perspective  \\


\bottomrule
\end{tabular}

\caption{}
\label{tbl:appendix:datapoints}
\end{table}

\end{document}